\documentclass[conference]{IEEEtran}
\IEEEoverridecommandlockouts

\usepackage[dvipsnames,table,xcdraw]{xcolor}

\usepackage[pagebackref=true,breaklinks=true,colorlinks,citecolor=ForestGreen]{hyperref}
\usepackage{cite}
\usepackage{amsmath,amssymb,amsfonts}
\usepackage{algorithmic}
\usepackage{graphicx}
\usepackage{textcomp}
\usepackage{xcolor}  
\usepackage{booktabs}
\usepackage{url}
\usepackage{tablefootnote}
\usepackage{hyperref}
\def\BibTeX{{\rm B\kern-.05em{\sc i\kern-.025em b}\kern-.08em
    T\kern-.1667em\lower.7ex\hbox{E}\kern-.125emX}}
\begin{document}

\title{RAWIC: Bit-Depth Adaptive Lossless Raw Image Compression
\thanks{$\star$ Work done during an internship at AIR, Tsinghua University}
\thanks{$^{\dagger }$ Corresponding authors.}}


\author{
\IEEEauthorblockN{Chunhang Zheng$^{1, \star}$, Tongda Xu$^{2}$, Mingli Xie$^{3}$, Yan Wang$^{2,\dagger}$, Dou Li$^{1,\dagger}$}
\IEEEauthorblockA{$^1$ School of Electronics, Peking University, Beijing, 100871, China}
\IEEEauthorblockA{$^2$ Institute for AI Industry Research (AIR), Tsinghua University, Beijing, 100084, China}
\IEEEauthorblockA{$^3$ the University of Chinese Academy of Sciences, Beijing, 100190, China}
 
\IEEEauthorblockA{zhengch@stu.pku.edu.cn, wangyan@air.tsinghua.edu.cn, lidou@pku.edu.cn}
}
\maketitle

\begin{abstract}
Raw images preserve linear sensor measurements and high bit-depth information crucial for advanced vision tasks and photography applications, yet their storage remains challenging due to large file sizes, varying bit depths, and sensor-dependent characteristics. Existing learned lossless compression methods mainly target 8-bit sRGB images, while raw reconstruction approaches are inherently lossy and rely on camera-specific assumptions.
To address these challenges, we introduce RAWIC, a bit-depth-adaptive learned lossless compression framework for Bayer-pattern raw images. We first convert single-channel Bayer data into a four-channel RGGB format and partition it into patches. For each patch, we compute its bit depth and use it as auxiliary input to guide compression. A bit-depth-adaptive entropy model is then designed to estimate patch distributions conditioned on their bit depths.
This architecture enables a single model to handle raw images from diverse cameras and bit depths. Experiments show that RAWIC consistently surpasses traditional lossless codecs, achieving an average 7.7\% bitrate reduction over JPEG-XL. Our code is available at \url{https://github.com/chunbaobao/RAWIC}.
\end{abstract}


\section{Introduction}\label{sec:introduction}

Raw images are unprocessed and uncompressed measurements directly captured by camera sensors. Owing to their linear relationship with scene radiance and their preservation of the original signal before any image signal processing (ISP), raw data have become indispensable for numerous low-level vision tasks, including image denoising~\cite{zhang2021rethinking}, image super-resolution~\cite{xu2019towards}, and low-light enhancement~\cite{huang2022towards}. Compared with processed standard RGB (sRGB) images rendered from raw image through ISP pipelines, raw images retain richer intensity information, higher bit depth, and sensor-specific characteristics that are crucial for downstream computational photography and imaging applications.

However, the storage and transmission of raw images pose significant challenges due to their (i) different
characteristics of various camera sensors
and (ii) variable bit depths, typically ranging from 10 to 14 bits. These factors result in substantial file sizes, making efficient compression techniques essential for practical deployment. To improve storage efficiency, a series of raw image reconstruction methods~\cite{punnappurath2021spatially, nam2022learning, wang2023raw, wang2024beyond} have been explored to recover high-quality raw data either blindly or with the assistance of limited metadata.


Despite their promising performance, reconstruction-based methods still suffer from several fundamental limitations. First, they are inherently lossy, as they attempt to approximate the original raw data from sRGB images or sparse samples, inevitably disrupting the linear radiance relationship that raw data are expected to preserve.
Second, these methods typically operate on three-channel representations after demosaicing rather than the original single-channel Bayer pattern measurements, introducing additional non-linear distortions.
Third, these methods are often camera-specific and bit-depth-specific, requiring separate models for different camera sensors and bit depths, which limits their generalizability and practical applicability.


\begin{figure}[t!]
    \centering
    \includegraphics[width=1\linewidth]
    {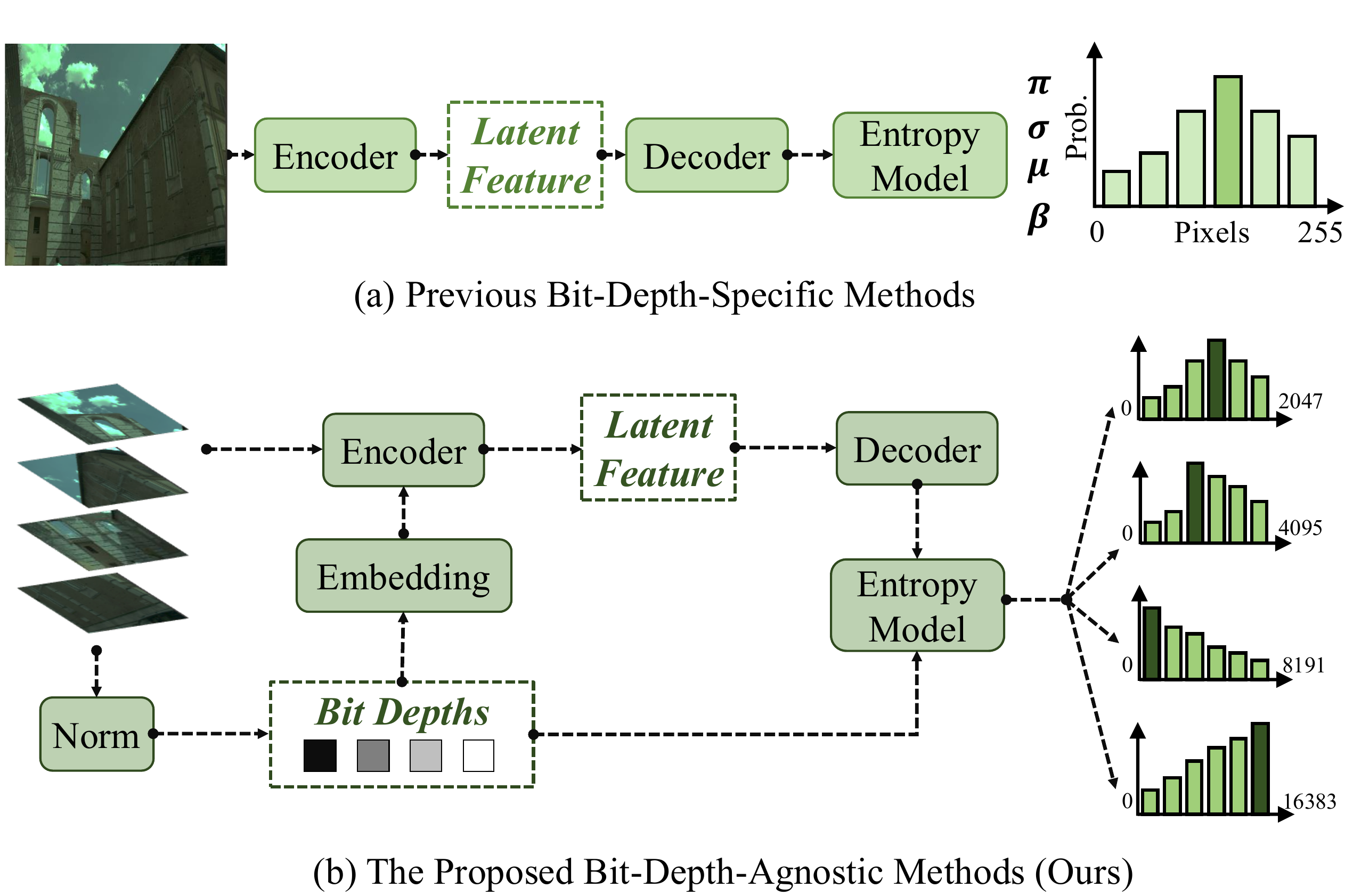}
    \caption{The comparison between previous bit-depth-specific lossless image compression methods and our proposed bit-depth adaptive lossless raw image compression method.
        (a) Exiting methods train the model a specific bit depth and can only compress raw images with the same bit depth. (b) Our proposed method uses a single model to compress raw images across different bit depths.}
    \label{fig:ep_comp}
\end{figure}

To address these limitations, instead reconstruction raw image from metadata, we propose RAWIC, a novel framework for bit-depth adaptive lossless raw image compression that can effectively accommodate different bit depths and cameras. Specifically, we first convert the single-channel Bayer pattern raw images from different cameras into an RGGB four-channel representation and split the resulting data into patches. For each patch, we compute its bit depth and feed this information into the model as an auxiliary input to guide the compression process. A bit-depth adaptive entropy model then estimates the probability distribution of raw patches conditioned on their bit depths, enabling efficient and fully lossless compression across a wide range of bit depths.
Figure~\ref{fig:ep_comp} illustrates the distinction between our proposed method and previous methods.

Our key contributions are summarized as follows:

\begin{itemize}
    \item We propose the first learned lossless raw image compression framework that directly compresses single-channel Bayer pattern raw images without any distortions introduced by demosaicing or lossy reconstruction.
    \item We design a camera-agnostic and bit-depth adaptive architecture that enables a single model to compress raw images from different camera sensors and bit depths, enhancing its versatility and practical utility.
    \item We extensively evaluate our proposed RAWIC framework on multiple raw image datasets with varying bit depths, demonstrating its superior compression performance compared to existing methods.
\end{itemize}
\section{Related Works}\label{sec:related-works}

\subsection{Raw Image Reconstruction}
Raw image reconstruction aims to recover raw images from their rendered sRGB counterparts. According if there is additional metadata available during reconstruction, existing methods can be broadly categorized into two groups: blind raw reconstruction and raw reconstruction with metadata.\\
\textbf{Blind Raw Reconstruction.} Blind raw reconstruction methods aim to recover raw images solely from sRGB inputs without relying on any auxiliary information. Since early digital cameras did not provide access to sensor-level raw RGB data, initial research in this area primarily focused on linearizing sRGB images, a problem commonly formulated as radiometric calibration~\cite{mitsunaga1999radiometric}. Subsequent works attempted to better describe the pipeline of ISP as access to raw sensor data became more prevalent~\cite{ chakrabarti2014modeling, gong2018rank}.\\
\textbf{Raw Reconstruction with Metadata.} With the increasing availability of raw images and their associated metadata, recent approaches have leveraged this additional information to enhance reconstruction quality. Metadata such as low resolution raw~\cite{yuan2011high}, estimated parameters of the simplified ISP~\cite{nguyen2016raw}, and sample of raw images~\cite{punnappurath2021spatially, nam2022learning} have been utilized to guide the reconstruction process. Most recently, Wang et al.~\cite{wang2023raw, wang2024beyond} proposed to treat raw reconstruction as a learned lossy compression problem, where sRGB images serve as priors and the resulting latent features act as metadata to facilitate more accurate raw recovery.

\subsection{Learned Lossless Image Compression}
Unlike learned lossy codecs that optimize the distortion metric between original and reconstructed images, lossless learned compression approaches focus on directly modeling the probability distributions of pixel intensities. These methods aim to minimize the total bit rate, which comprises contributions from both latent representations and pixel data:
\begin{equation}\label{eq:tworate}
    \mathcal{L}=\mathcal{R}=\mathcal{R}_\text{latent}+\mathcal{R}_\text{pixel},
\end{equation}

The efficacy of learned lossless compression hinges on how precisely the pixel probability distributions are estimated. To capture inter-pixel dependencies effectively, researchers have developed diverse neural architectures, which can be grouped into four main categories: autoregressive approaches, variational autoencoder (VAE) frameworks, normalizing flow methods, and overfitting-based methods.
Autoregressive approaches, exemplified by PixelCNN~\cite{van2016conditional} and PixelCNN++~\cite{salimans2017pixelcnn++}, employ recurrent or convolutional architectures to factorize the joint pixel distribution as $p(x_1, x_2, \ldots, x_n) = \prod_{i=1}^n p(x_i | x_1, x_2, \ldots, x_{i-1})$. VAE-based methods, including PixelVAE~\cite{gulrajani2016pixelvae}, DLPR~\cite{bai2024deep} and SEEC~\cite{zheng2025seec}, model pixel probabilities conditional on learned latent representations by optimizing the evidence lower bound (ELBO) through variational autoencoders~\cite{kingma2013auto}. Normalizing flow techniques, such as iFlow~\cite{zhang2021iflow}, leverage invertible transformations to map simple base distributions to complex data distributions through differentiable bijective mappings. 
Overfitting-based methods, like CALLIC~\cite{li2025callic} and FNLIC~\cite{zhang2025fitted}, achieve lossless compression by training models specifically on individual images, and storing the model parameters alongside the compressed data.
These methods each exhibit their own strengths and limitations with respect to compression ratio and computational efficiency.

\section{Methodology}\label{sec:methodologies}

\subsection{Motivation}

\begin{figure}[t!]
    \centering
    \includegraphics[width=1\linewidth]
    {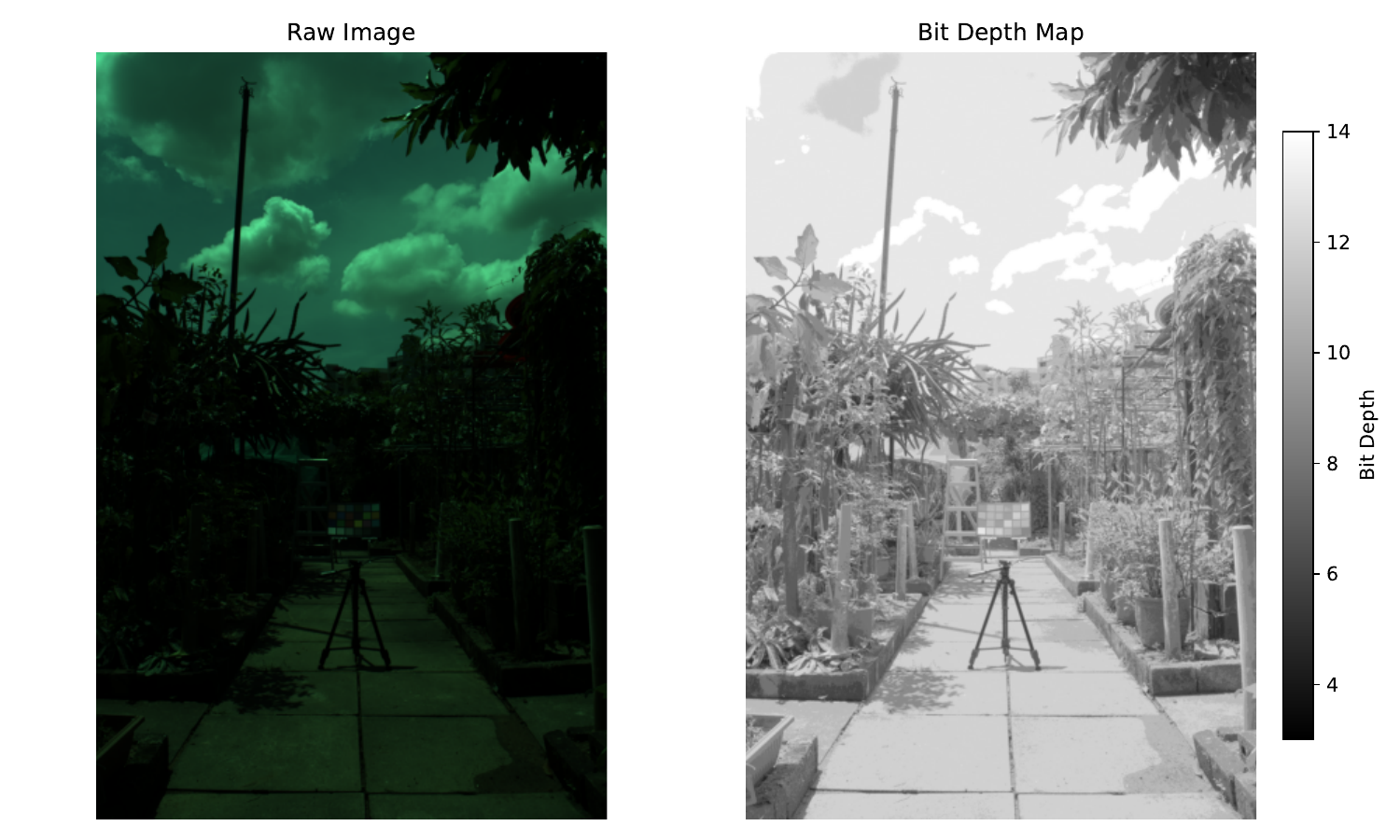}
    \caption{Visualization of raw image and corresponding pixel-wise bit depth map. The pixel-wise bit depth is computed as $\lceil \log_2(\text{value}+1) \rceil$. It can be observed that different regions in the raw image exhibit varying bit depths.}
    \label{fig:bit_depth_visualization}
\end{figure}

Existing learned lossless image compression methods predominantly focus on sRGB images with a fixed 8-bit depth per channel. However, raw images captured by different camera sensors exhibit varying bit depths, typically ranging from 10 to 14 bits. Moreover, even within a single raw image, pixel values can vary substantially as shown in Figure~\ref{fig:bit_depth_visualization}; thus, using a fixed bit depth for the entire image inevitably leads to significant bit redundancy.

\begin{figure*}[!t]
    \centering
    \includegraphics[width=1\linewidth]
    {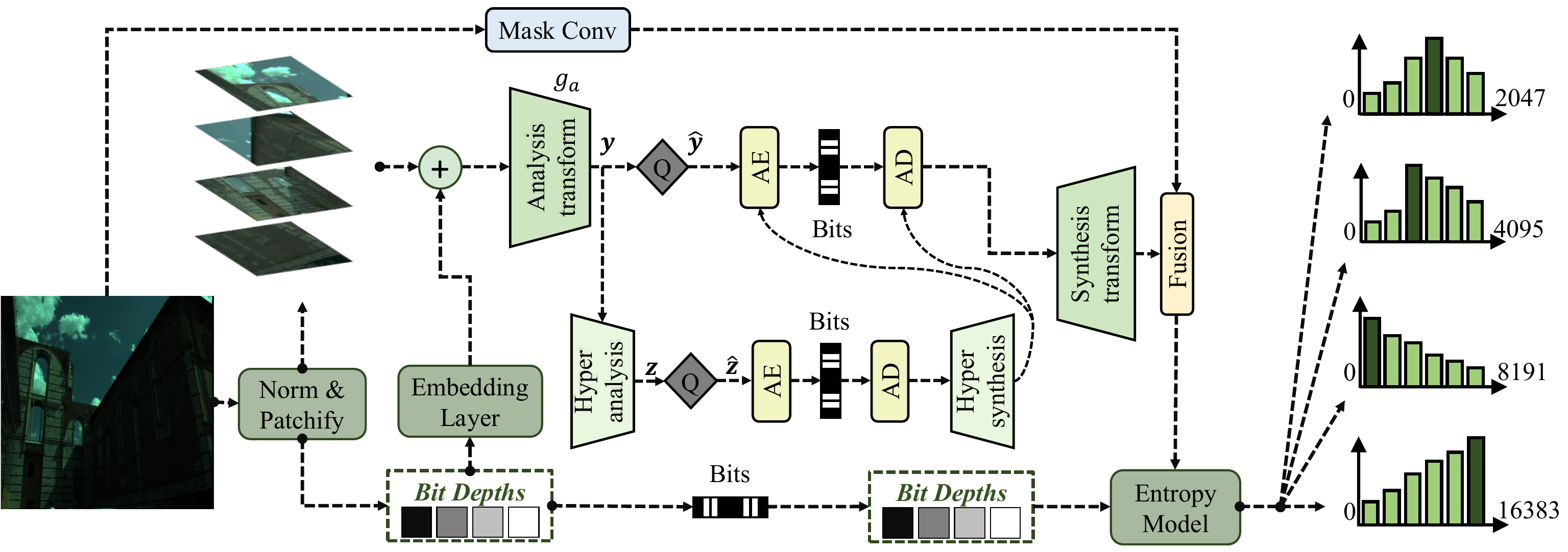}
    \caption{The overall architecture of RAWIC. 
        The input raw image is first converted to an RGGB four-channel format and split into patches. Each patch is then processed by a bit-depth adaptive entropy model that estimates the pixel distribution conditioned on the bit depth of the patch. Q denotes the quantization operation. AE and AD represent arithmetic encoder and arithmetic decoder, respectively.}
    \label{fig:pipeline}
\end{figure*}

\subsection{The overall Framework}
To tackle the challenges posed by varying bit depths in raw images, we propose RAWIC, a bit-depth adaptive lossless raw image compression framework, as illustrated in Figure~\ref{fig:pipeline}. Starting with a Bayer pattern raw image $\boldsymbol{x}$ containing a single channel, we transform it into a four-channel RGGB format and partition it into $N$ non-overlapping patches denoted as $\{\boldsymbol{x}_i\}_{i=1}^N$. For each patch $\boldsymbol{x}_i$, we compute its bit depth $\boldsymbol{b}_i$. 


The bit depth $\boldsymbol{b}_i$ is then fed into embedding layers to obtain the corresponding bit depth embedding $\boldsymbol{e}_i$.
We employ hyper-prior~\cite{balle2018variational} based ELIC framework~\cite{he2022elic} as our backbone architecture. Specifically, the input patch $\boldsymbol{x}_i$ is first transformed into a latent representation using an analysis transform conditioned on the bit depth embedding $\boldsymbol{y}_i = g_a(\boldsymbol{x}_i, \boldsymbol{e}_i)$.
Subsequently, the quantized latent is then passed through a hyper analysis to generate side information $\boldsymbol{z}_i = h_a(\boldsymbol{\hat{y}}_i)$, which is used to estimate the distribution parameters of the quantized latent representation $\boldsymbol{\hat{y}}_i$. We model the quantized latent $\boldsymbol{\hat{y}}_i$ using a Gaussian distribution, where the mean $\boldsymbol{\mu}_i$ and variance $\boldsymbol{\sigma}_i$ are predicted via hyper synthesis: $\boldsymbol{\mu}_i, \boldsymbol{\sigma}_i = h_s(\boldsymbol{\hat{z}}_i)$. The prior features $\boldsymbol{f}_i$ are derived from the quantized latent through synthesis transformation $\boldsymbol{f}_i = g_s(\boldsymbol{\hat{y}}_i)$. To enable end-to-end training, quantization is simulated by incorporating uniform noise: $p(\boldsymbol{\tilde{y}}_i) = \prod_{j} \mathcal{U}(\boldsymbol{y}_{ij} - \frac{1}{2}, \boldsymbol{y}_{ij} + \frac{1}{2})$ and $p(\boldsymbol{\tilde{z}}_i) = \prod_{j} \mathcal{U}(\boldsymbol{z}_{ij} - \frac{1}{2}, \boldsymbol{z}_{ij} + \frac{1}{2})$, where $j$ indexes the latent variables within the $i$-th patch.

Leveraging the prior features $\boldsymbol{f}_i$ alongside context features $\boldsymbol{C}_{\boldsymbol{x}_i}$ obtained from causal convolutions, we derive the distribution for individual pixels within raw patch $\boldsymbol{x}_i$:
\begin{equation}
    p(\boldsymbol{x}_i \mid \boldsymbol{C}_{\boldsymbol{x}_i}, \boldsymbol{f}_i)= p_{\theta} (\boldsymbol{x}_i \mid \boldsymbol{\mathcal{P}}_i)=
    \prod_{j=1}^{H \times W} p(\boldsymbol{x}_{ij} \mid \boldsymbol{x}_{i,<j}, \boldsymbol{f}_i)  , \label{ep:thetame}
\end{equation}
Where the $\theta$ denotes the parameters of the entropy model, and $\boldsymbol{\mathcal{P}}_i$ denotes the prior for the patch $\boldsymbol{x}_i$.

\begin{table*}[t!]
  \centering
  \caption{Lossless image compression performance (bpp) of our proposed method compared to other lossless image codec Canon 1Ds MkIII, Canon 600D, Olympus EPL6, Panasonic GX1, Samsung NX2000 and RAISE. The best results are highlighted in \textbf{bold}. We show the difference (\%) between our method and each traditional codec in \textcolor[HTML]{009900}{green} color. Note that our method includes the bitrates for storing bit depths.}
  \label{tab:bpp_results}
  \renewcommand{\arraystretch}{1.35}
  \setlength{\tabcolsep}{6pt}
  \begin{tabular*}{1\textwidth}{@{\extracolsep{\fill}}cccccccc@{}}
    \toprule
    Codec & Canon 1Ds MkIII & Canon 600D & Olympus EPL6 & Panasonic GX1 & Samsung NX2000 & RAISE \\
    \midrule
    QOI~\cite{qoi} 
      & 11.36 \scriptsize\textcolor[HTML]{009900}{(+67.3\%)} 
      & 12.09 \scriptsize\textcolor[HTML]{009900}{(+61.9\%)} 
      & 10.22 \scriptsize\textcolor[HTML]{009900}{(+100.0\%)} 
      & 10.97 \scriptsize\textcolor[HTML]{009900}{(+83.1\%)} 
      & 10.73 \scriptsize\textcolor[HTML]{009900}{(+84.1\%)} 
      & 12.86 \scriptsize\textcolor[HTML]{009900}{(+64.9\%)} \\

    PNG~\cite{boutell1997png}
      & 10.28 \scriptsize\textcolor[HTML]{009900}{(+51.4\%)} 
      & 10.96 \scriptsize\textcolor[HTML]{009900}{(+46.7\%)} 
      & 8.13 \scriptsize\textcolor[HTML]{009900}{(+59.1\%)} 
      & 9.45 \scriptsize\textcolor[HTML]{009900}{(+57.8\%)} 
      & 9.16 \scriptsize\textcolor[HTML]{009900}{(+57.1\%)} 
      & 11.37 \scriptsize\textcolor[HTML]{009900}{(+45.8\%)} \\

    WebP~\cite{webp}
      & 7.80 \scriptsize\textcolor[HTML]{009900}{(+14.9\%)} 
      & 8.44 \scriptsize\textcolor[HTML]{009900}{(+13.0\%)} 
      & 5.88 \scriptsize\textcolor[HTML]{009900}{(+15.1\%)} 
      & 7.07 \scriptsize\textcolor[HTML]{009900}{(+18.0\%)} 
      & 6.66 \scriptsize\textcolor[HTML]{009900}{(+14.2\%)} 
      & 8.82 \scriptsize\textcolor[HTML]{009900}{(+13.1\%)} \\

    FLIF~\cite{sneyers2016flif}
      & 7.82 \scriptsize\textcolor[HTML]{009900}{(+15.2\%)} 
      & 8.34 \scriptsize\textcolor[HTML]{009900}{(+11.7\%)} 
      & 6.12 \scriptsize\textcolor[HTML]{009900}{(+19.8\%)} 
      & 7.02 \scriptsize\textcolor[HTML]{009900}{(+17.2\%)} 
      & 6.86 \scriptsize\textcolor[HTML]{009900}{(+17.7\%)} 
      & 8.86 \scriptsize\textcolor[HTML]{009900}{(+13.6\%)} \\

    JPEG2000~\cite{skodras2001jpeg}
      & 7.34 \scriptsize\textcolor[HTML]{009900}{(+8.1\%)} 
      & 8.04 \scriptsize\textcolor[HTML]{009900}{(+7.7\%)} 
      & 5.70 \scriptsize\textcolor[HTML]{009900}{(+11.6\%)} 
      & 6.67 \scriptsize\textcolor[HTML]{009900}{(+11.3\%)} 
      & 6.58 \scriptsize\textcolor[HTML]{009900}{(+12.9\%)} 
      & 8.58 \scriptsize\textcolor[HTML]{009900}{(+10.0\%)} \\

    JPEG-LS~\cite{weinberger2000loco}
      & 7.25 \scriptsize\textcolor[HTML]{009900}{(+6.8\%)} 
      & 7.87 \scriptsize\textcolor[HTML]{009900}{(+5.4\%)} 
      & 5.67 \scriptsize\textcolor[HTML]{009900}{(+11.0\%)} 
      & 6.54 \scriptsize\textcolor[HTML]{009900}{(+9.2\%)} 
      & 6.49 \scriptsize\textcolor[HTML]{009900}{(+11.3\%)} 
      & 8.38 \scriptsize\textcolor[HTML]{009900}{(+7.4\%)} \\

    JPEG-XL~\cite{alakuijala2019jpeg}
      & 7.29 \scriptsize\textcolor[HTML]{009900}{(+7.4\%)} 
      & 7.95 \scriptsize\textcolor[HTML]{009900}{(+6.4\%)} 
      & 5.60 \scriptsize\textcolor[HTML]{009900}{(+9.6\%)} 
      & 6.57 \scriptsize\textcolor[HTML]{009900}{(+9.7\%)} 
      & 6.46 \scriptsize\textcolor[HTML]{009900}{(+10.8\%)} 
      & 8.29 \scriptsize\textcolor[HTML]{009900}{(+6.3\%)} \\

    \midrule
    RAWIC (Ours)
      & \textbf{6.79} & \textbf{7.47} & \textbf{5.11} & \textbf{5.99} & \textbf{5.83} & \textbf{7.80} \\
    \bottomrule
  \end{tabular*}
\end{table*}

\subsection{Bit-Depth Adaptive Entropy Model}

Fixed bit-depth entropy models are suboptimal for raw images due to significant variations in pixel values both across different images and within a single image. To address this challenge, we introduce a bit-depth adaptive entropy model that dynamically adapts to the varying bit depths of raw image patches.

Given a pixel $\boldsymbol{x}_{ij}^{c}$ (where $c$ denotes the channel index) with predicted probability mass function (PMF) $p(\boldsymbol{x}_{ij}^{c})$ produced by the bit-depth-specific entropy model for patch $\boldsymbol{x}_i$ and its corresponding bit depth $\boldsymbol{b}_{ij}^{c}$, we define the bit-depth adaptive probability distribution as follows:
\begin{equation}
    \tilde{p}(\boldsymbol{x}_{ij}^{c} \mid \boldsymbol{b}_{ij}^{c}) = \frac{p(\boldsymbol{x}_{ij}^{c}) \cdot \mathbb{I}(0 \leq \boldsymbol{x}_{ij}^{c} < 2^{\boldsymbol{b}_{ij}^{c}})}{\sum_{k=0}^{2^{\boldsymbol{b}_{ij}^{c}}-1} p(k)}, \label{eq:bit_depth_adaptive}
\end{equation}
where $\mathbb{I}(\cdot)$ is an indicator function that masks out invalid pixel values outside the valid range $[0, 2^{\boldsymbol{b}_{ij}^{c}}-1]$, and the denominator renormalizes the PMF to ensure $\sum_{k=0}^{2^{\boldsymbol{b}_{ij}^{c}}-1} \tilde{p}(k \mid \boldsymbol{b}_{ij}^{c}) = 1$.


To model the distribution of pixel values in raw images with RGGB Bayer pattern, we extend the discrete logistic mixture likelihood~\cite{salimans2017pixelcnn++} to handle multi-channel raw image data. For a raw image patch $\boldsymbol{x}_i$ in RGGB format, the likelihood factorizes as:
\begin{equation}
    p_{\theta}(\boldsymbol{x}_i \mid \boldsymbol{\mathcal{P}}_i) = \prod_{j=1}^{H \times W} p_{\theta}(\boldsymbol{x}_{ij}^{\text{r}}, \boldsymbol{x}_{ij}^{\text{g}_1}, \boldsymbol{x}_{ij}^{\text{g}_2}, \boldsymbol{x}_{ij}^{\text{b}} \mid \boldsymbol{\mathcal{P}}_i) , \label{eq:likelihood}
\end{equation} 
where $\boldsymbol{x}_{ij}^{\text{r}}$, $\boldsymbol{x}_{ij}^{\text{g}_1}$, $\boldsymbol{x}_{ij}^{\text{g}_2}$, and $\boldsymbol{x}_{ij}^{\text{b}}$ represent the pixel values at spatial location $j$ in the red, first green, second green, and blue channels of the Bayer pattern, respectively. Channel autoregressive modeling~\cite{salimans2017pixelcnn++} is employed to capture inter-channel dependencies, factorizing the joint distribution as:
\begin{align}
&p_{\theta}(\boldsymbol{x}_{ij}^{\text{r}}, \boldsymbol{x}_{ij}^{\text{g}_1}, 
\boldsymbol{x}_{ij}^{\text{g}_2}, \boldsymbol{x}_{ij}^{\text{b}} \mid \boldsymbol{\mathcal{P}}_i)
= p_{\theta}(\boldsymbol{x}_{ij}^{\text{r}} \mid \boldsymbol{\mathcal{P}}_i)
   \cdot p_{\theta}(\boldsymbol{x}_{ij}^{\text{g}_1} \mid \boldsymbol{x}_{ij}^{\text{r}}, \boldsymbol{\mathcal{P}}_i) \nonumber \\
&\cdot p_{\theta}(\boldsymbol{x}_{ij}^{\text{g}_2} \mid \boldsymbol{x}_{ij}^{\text{r}}, 
   \boldsymbol{x}_{ij}^{\text{g}_1}, \boldsymbol{\mathcal{P}}_i)
   \cdot p_{\theta}(\boldsymbol{x}_{ij}^{\text{b}} \mid \boldsymbol{x}_{ij}^{\text{r}}, 
   \boldsymbol{x}_{ij}^{\text{g}_1}, \boldsymbol{x}_{ij}^{\text{g}_2}, \boldsymbol{\mathcal{P}}_i).
\label{eq:autoregressive}
\end{align}

Each conditional distribution is modeled as a discrete logistic mixture with $K$ components. For channel $c \in \{\text{r}, \text{g}_1, \text{g}_2, \text{b}\}$, the probability mass function is given by:
\begin{align}
p_{\theta}(\boldsymbol{x}_{ij}^{c} \mid \cdot)
&= \sum_{k=1}^{K} \boldsymbol{\pi}_{ij}^{ck} \cdot 
\Bigg[
    \sigma\!\left(\frac{\boldsymbol{x}_{ij}^{c} + \boldsymbol{\Delta}_{ij}^{c}/2 - \boldsymbol{\mu}_{ij}^{ck}}{\boldsymbol{s}_{ij}^{ck}}\right)
    - \nonumber
\\
&\qquad\qquad
    \sigma\!\left(\frac{\boldsymbol{x}_{ij}^{c} - \boldsymbol{\Delta}_{ij}^{c}/2 - \boldsymbol{\mu}_{ij}^{ck}}{\boldsymbol{s}_{ij}^{ck}}\right)
\Bigg].
\label{eq:mixture}
\end{align}
where $\sigma(\cdot)$ is the sigmoid function, $\boldsymbol{\mu}_{ij}^{ck}$ and $\boldsymbol{s}_{ij}^{ck}$ are the mean and scale of the $k$-th mixture component, $\boldsymbol{\pi}_{ij}^{ck}$ is the mixture weight with $\sum_{k=1}^{K} \boldsymbol{\pi}_{ij}^{ck} = 1$, and $\boldsymbol{\Delta}_{ij}^{c}$ represents the quantization bin width determined by the bit depth $\boldsymbol{b}_{ij}^{c}$ as $\boldsymbol{\Delta}_{ij}^{c} = 1 / (2^{\boldsymbol{b}_{ij}^{c}} - 1)$. For edge cases where $\boldsymbol{x}_{ij}^{c}=0$ or $\boldsymbol{x}_{ij}^{c}=2^{\boldsymbol{b}_{ij}^{c}}-1$, $\boldsymbol{x}_{ij}^{c} - \boldsymbol{\Delta}_{ij}^{c}/2$ or $\boldsymbol{x}_{ij}^{c} + \boldsymbol{\Delta}_{ij}^{c}/2$ is clipped to $-\infty$ or $\infty$, respectively.

The autoregressive dependencies are modeled through the means of the mixture components:
\begin{align}
    &\boldsymbol{\hat{\mu}}_{ij}^{\text{r}k} = \boldsymbol{\mu}_{ij}^{\text{r}k}, \nonumber 
    \quad \boldsymbol{\hat{\mu}}_{ij}^{\text{g}_1k} = \boldsymbol{\mu}_{ij}^{\text{g}_1k} + \boldsymbol{\beta}_{ij}^{\text{r}k} \cdot {\boldsymbol{x}}_{ij}^{\text{r}},  \nonumber  \\
    &\boldsymbol{\hat{\mu}}_{ij}^{\text{g}_2k} = \boldsymbol{\mu}_{ij}^{\text{g}_2k} + \boldsymbol{\beta}_{ij}^{\text{g}_1k} \cdot {\boldsymbol{x}}_{ij}^{\text{g}_1}, \nonumber \\
    &\boldsymbol{\hat{\mu}}_{ij}^{\text{b}k} = \boldsymbol{\mu}_{ij}^{\text{b}k} + \boldsymbol{\beta}_{ij}^{\text{g}_2k} \cdot {\boldsymbol{x}}_{ij}^{\text{r}} + \boldsymbol{\beta}_{ij}^{bk} \cdot \left({\boldsymbol{x}}_{ij}^{\text{g}_1} + {\boldsymbol{x}}_{ij}^{\text{g}_2}\right) / 2, \label{eq:mu_b}
\end{align}
where $\boldsymbol{\beta}_{ij}^{ck}$ are learnable coefficients that capture inter-channel dependencies. 

All distribution parameters, including the base means $\boldsymbol{\mu}_{ij}^{ck}$, scales $\boldsymbol{s}_{ij}^{ck}$, mixture weights $\boldsymbol{\pi}_{ij}^{ck}$, and autoregressive coefficients $\boldsymbol{\beta}_{ij}^{ck}$, are jointly produced by the entropy parameter network conditioned on the prior $\boldsymbol{\mathcal{P}}_i$ and bit depth $\boldsymbol{b}_{ij}^{c}$.


\section{Experiments}\label{sec:experiments}

\subsection{Settings}\label{subsec:settings}

We train our RAWIC model using raw data from five cameras in the NUS dataset~\cite{cheng2014illuminant}, including Canon 1Ds MkIII, Canon 600D, Olympus EPL6, Panasonic GX1, and Samsung NX2000, whose raw images have bit depths of 14, 14, 12, 12, 12  and 14 bits, respectively.
For each camera, we randomly split the images into 80\% for training, 10\% for validation, and 10\% for testing. In addition, we incorporate 5\% of randomly selected raw images from the RAISE dataset~\cite{dang2015raise}. In total, 1088 raw images are used for training.
During training, raw images are initially split into non-overlapping $128 \times 128$ patches. We apply horizontal and vertical flipping with a probability of $0.5$, then randomly crop the patches to $64 \times 64$. The model is trained for $200$ epochs using the Adam optimizer~\cite{kingma2014adam} with a batch size of $128$. The learning rate is initialized at $1 \times 10^{-4}$ and adaptively reduced by a factor of $0.1$ if the validation loss does not improve for $10$ consecutive epoch.
The implementation of the model is based on CompressAI~\cite{begaint2020compressai}. We train the RAWIC model on NVIDIA A100 GPU.

\subsection{Coding Performance}

To evaluate the effectiveness of our proposed RAWIC codec, we compare it against seven traditional lossless image compression methods, including QOI~\cite{qoi}, PNG~\cite{boutell1997png}, JPEG-LS~\cite{weinberger2000loco}, JPEG2000~\cite{skodras2001jpeg}, WebP~\cite{webp}, JPEG-XL~\cite{alakuijala2019jpeg}, and FLIF~\cite{sneyers2016flif}. We split each raw image into a most-significant-byte (MSB) subimage and a least-significant-byte (LSB) subimage when evaluating WebP and FLIF, since both formats only support 8-bit-per-channel images.
All methods are evaluated in terms of bits per pixel (bpp) on the corresponding test sets, as shown in Table~\ref{tab:bpp_results}.
Our RAWIC model consistently outperforms all traditional codecs across different camera sensors and bit depth with one single trained model.
On average, our method achieves a bitrate reduction of 7.7\% compared to the best traditional method, JPEG-XL.

To compare with sRGB lossless image compression methods, we also extend our RAWIC to compress sRGB images with 8-bit per channel.
Our model is retrained on the DIV2K dataset and evaluated alongside a range of recent learned lossless compression methods including L3C~\cite{mentzer2019practical},
RC~\cite{mentzer2020learning}, SReC~\cite{cao2020lossless},
iVPF~\cite{zhang2021ivpf}, iFlow~\cite{zhang2021iflow},
Near-Lossless~\cite{bai2021learning}, ArIB-BPS~\cite{zhang2024learned}, and DLPR~\cite{bai2024deep}. We evaluate all methods on the DIV2K validation set, CLIC~\cite{clic2020dataset} and the Kodak~\cite{kodak}. For fair comparison, we exclude LLM-based
methods and overfitting-based methods due to their excessive
encoding or decoding time. The results are summarized in Table~\ref{tab:rgb_bpp_results}.

\begin{table}[htbp]
  \centering
  \renewcommand{\arraystretch}{1.3}
  \setlength{\tabcolsep}{8pt}
  \caption{Lossless RGB image compression performance (bpp) compared to other learned lossless image codecs.}
  \label{tab:rgb_bpp_results}
  \begin{tabular}{lccc}
    \toprule
    Codec & DIV2K & CLIC.m & Kodak  \\
    \midrule
    L3C~\cite{mentzer2019practical} 
      & 9.27 \scriptsize\textcolor[HTML]{009900}{(+22.9\%)} 
      & 7.92 \scriptsize\textcolor[HTML]{009900}{(+23.4\%)} 
      & 9.78 \scriptsize\textcolor[HTML]{009900}{(+15.5\%)} \\

    RC~\cite{mentzer2020learning} 
      & 9.24 \scriptsize\textcolor[HTML]{009900}{(+22.6\%)} 
      & 7.62 \scriptsize\textcolor[HTML]{009900}{(+18.7\%)} 
      & - \\

    SReC~\cite{cao2020lossless} 
      & 8.47 \scriptsize\textcolor[HTML]{009900}{(+12.3\%)} 
          & 7.32 \scriptsize\textcolor[HTML]{009900}{(+14.0\%)} 
      & 9.10 \scriptsize\textcolor[HTML]{009900}{(+7.4\%)} \\

    Near-Lossless~\cite{bai2021learning} 
      & 8.43 \scriptsize\textcolor[HTML]{009900}{(+11.8\%)} 
      & 7.53 \scriptsize\textcolor[HTML]{009900}{(+17.3\%)} 
      & 9.12 \scriptsize\textcolor[HTML]{009900}{(+7.7\%)} \\

    iVPF~\cite{zhang2021ivpf} 
      & 8.04 \scriptsize\textcolor[HTML]{009900}{(+6.6\%)} 
      & 7.17 \scriptsize\textcolor[HTML]{009900}{(+11.7\%)} 
      & - \\

    iFlow~\cite{zhang2021iflow} 
      & 7.71 \scriptsize\textcolor[HTML]{009900}{(+2.3\%)} 
      & 6.78 \scriptsize\textcolor[HTML]{009900}{(+5.6\%)} 
      & - \\

    ArIB-BPS~\cite{zhang2024learned} 
      & 7.65 \scriptsize\textcolor[HTML]{009900}{(+1.5\%)} 
      & - 
      & - \\

    DLPR~\cite{bai2024deep} 
      & 7.65 \scriptsize\textcolor[HTML]{009900}{(+1.5\%)} 
      & 6.48 \scriptsize\textcolor[HTML]{009900}{(+0.9\%)} 
      & 8.58 \scriptsize\textcolor[HTML]{009900}{(+1.3\%)} \\

    \midrule
    RAWIC (Ours) 
      & \textbf{7.54} 
      & \textbf{6.42} 
      & \textbf{8.47} \\
    \bottomrule
  \end{tabular}
\end{table}

\subsection{Efficiency of Codec}

To evaluate the efficiency of our RAWIC codec, we measure the average runtime (in seconds) required to encode and decode a single raw image. We compare the runtime of our RAWIC framework with several traditional lossless image codecs on raw images from the Canon 600D and Olympus EPL6 datasets, whose raw resolutions are $3464 \times 5202$ and $3472 \times 4640$ with bit depths of 14 bits and 12 bits, respectively. The results are summarized in Table~\ref{tab:efficiency}.

\begin{table}[htbp]
  \centering
  \renewcommand{\arraystretch}{1.2}
  \setlength{\tabcolsep}{8pt}
  \caption{Average runtime (sec.) of RAWIC compared to other codecs (Encoding/Decoding time).}

  \begin{tabular}{lcccc}
    \toprule
    Codec                             &    Canon 600D  &   Olympus EPL6   &   \\         
    \midrule
    PNG~\cite{boutell1997png}        &  4.55/0.59    & 4.56/0.46     \\
    WebP~\cite{webp}        &  5.99/0.26     & 3.86/0.18      \\
    JPEG-XL~\cite{alakuijala2019jpeg}   &  6.08/1.38    & 6.62/1.46     \\
    FLIF~\cite{sneyers2016flif}       & 25.97/9.36      & 20.41/9.36      \\

    \midrule
    RAWIC (Ours)                      &  45.7/119.3   & 37.1/98.4 \\

    \bottomrule
  \end{tabular}
  \label{tab:efficiency}
\end{table}

We also evaluate the runtime of our RAWIC codec on sRGB images with 8-bit per channel. We compare the runtime of our RAWIC framework with several learned lossless image codecs on RGB images from the Kodak dataset with different resolutions. The results are summarized in Table~\ref{tab:rgb_efficiency}.

\begin{table}[htbp]
  \centering
  \renewcommand{\arraystretch}{1.3}
  \setlength{\tabcolsep}{10pt}
  \caption{Average runtime (sec.) of RAWIC compared to other codecs (Encoding/Decoding time). OOM: out of memory.}

  \begin{tabular}{lcccc}
  \toprule
  Codec & 768$\times$512 & 1024$\times$768  &  2048$\times$1536\\
  \midrule
WebP~\cite{webp} & 0.15/0.01 & 0.20/0.01 & 2.23/0.05 \\
JPEG-XL~\cite{alakuijala2019jpeg} & 0.24/0.06 & 0.39/0.09 & 2.97/0.53 \\
BPG~\cite{bellard2015bpg} & 0.23/0.19 & 0.44/0.32 & 5.71/1.42 \\
FLIF~\cite{sneyers2016flif} & 4.13/0.66 & 7.31/1.13 & 33.07/5.27 \\

SReC~\cite{cao2020lossless} & 0.58/0.59 & 1.10/1.16 & 4.11/4.58 \\

L3C~\cite{mentzer2019practical} & 0.70/0.64 & 1.30/1.26 & 4.93/5.06 \\
Minnen~\cite{minnen2018joint} & 1.26/2.50 & 2.48/5.01  & 9.97/19.81  \\
ArIB-BPS~\cite{zhang2024learned} & 6.59/6.42 & OOM & OOM
 \\
  DLPR ~\cite{bai2024deep} & 0.97/1.49 & 1.68 /2.76 & 6.96 /11.27 \\
  \midrule
  RAWIC (Ours) & 0.96/1.43 & 1.62/2.56 & 6.71/10.72 \\
  \bottomrule

\end{tabular}
\label{tab:rgb_efficiency}
\end{table}

\subsection{Ablation Study}

\subsubsection{Ablation of Bit-Depth Adaptive Entropy Model}

To validate the effectiveness of our proposed bit-depth adaptive entropy model, we conduct an ablation study by replacing it with a fixed-bit-depth entropy model that applies a constant bit depth to all raw patches. As shown in Table~\ref{tab:ablation_bde}, the bit-depth adaptive entropy model consistently outperforms the fixed-bit-depth counterpart across all camera sensors, with notably larger gains on datasets with lower bit depths. This demonstrates its effectiveness in modeling the varying bit-depth characteristics of raw images.

\begin{table}[htbp]
  \centering
  \renewcommand{\arraystretch}{1.2}
  \setlength{\tabcolsep}{8pt}
  \caption{Ablation study on the effectiveness of the bit-depth adaptive entropy model.}
  \label{tab:ablation_bde}
  \begin{tabular}{lcc}
    \toprule
    Dataset & Bit-Depth Adaptive (Ours) & Fixed Bit Depth \\
    \midrule
    Canon 1Ds MkIII & 6.79 & 8.78 \scriptsize\textcolor[HTML]{009900}{(+29.3\%)} \\
    Canon 600D & 7.47 & 8.77 \scriptsize\textcolor[HTML]{009900}{(+17.4\%)} \\
    Olympus EPL6 & 5.11 & 8.98 \scriptsize\textcolor[HTML]{009900}{(+75.7\%)} \\
    Panasonic GX1 & 5.99 & 9.20 \scriptsize\textcolor[HTML]{009900}{(+53.6\%)} \\
    Samsung NX2000 & 5.83 & 9.37 \scriptsize\textcolor[HTML]{009900}{(+60.7\%)} \\
    RAISE & 7.80 & 9.29 \scriptsize\textcolor[HTML]{009900}{(+19.1\%)} \\
    \bottomrule
  \end{tabular}
\end{table}

\subsubsection{Ablation of All-in-One Model}

To assess the advantage of our all-in-one model that compresses raw images across different camera sensors, we compare it with camera-specific models that are individually trained for each camera sensor. As shown in Table~\ref{tab:ablation_all_in_one}, our all-in-one model achieves better compression performance than camera-specific models on most datasets, demonstrating its effectiveness in handling raw images from diverse cameras with varying characteristics.

\begin{table}[htbp]
  \centering
  \renewcommand{\arraystretch}{1.2}
  \setlength{\tabcolsep}{6pt}
  \caption{Ablation study on the effectiveness of the all-in-one model.}
  \label{tab:ablation_all_in_one}
  \begin{tabular}{lcc}
    \toprule
    Dataset & All-in-One Model (Ours) & Camera-Specific Model \\
    \midrule
    Canon 1Ds MkIII & 6.79 & 7.91 \scriptsize\textcolor[HTML]{009900}{(+16.5\%)} \\
    Canon 600D      & 7.47 & 8.85 \scriptsize\textcolor[HTML]{009900}{(+18.5\%)} \\
    Olympus EPL6    & 5.11 & 6.79 \scriptsize\textcolor[HTML]{009900}{(+32.9\%)} \\
    Panasonic GX1   & 5.99 & 6.46 \scriptsize\textcolor[HTML]{009900}{(+7.8\%)} \\
    Samsung NX2000 & 5.83 & 5.82 \scriptsize\textcolor[HTML]{DC143C}{(-0.1\%)} \\
    RAISE          & 7.79 & 7.80 \scriptsize\textcolor[HTML]{009900}{(+0.1\%)} \\
    \bottomrule
  \end{tabular}
\end{table}


\section{Conclusion}\label{sec:conclusion}
In this paper, we propose RAWIC, a novel framework for bit-depth adaptive lossless raw image compression. By leveraging adaptive entropy modeling conditioned on bit depth, our method efficiently compresses raw images from diverse cameras and bit depths using a single unified model. Extensive experiments show that RAWIC consistently outperforms traditional lossless codecs. Future work will explore reducing the computational complexity and latency of the raw neural codec to enable real-time applications.
\section{ACKNOWLEDGEMENTS}




This work was supported by the National Key R\&D Program of China under Award Numbers 2020YFB1807802 and 2016ZX03001018-005, and by the Wuxi Research Institute of Applied Technologies, Tsinghua University under Grant 20242001120.

\bibliographystyle{IEEEbib}
\bibliography{icme2026references}

\end{document}